\documentclass{article}

\pdfoutput=1

\usepackage{arxiv}

\usepackage[utf8]{inputenc} 
\usepackage[T1]{fontenc}    
\usepackage{hyperref}       
\usepackage{url}            
\usepackage{booktabs}       
\usepackage{amsfonts}       
\usepackage{nicefrac}       
\usepackage{microtype}      
\usepackage{lipsum}		
\usepackage{graphicx}
\usepackage[numbers]{natbib}
\usepackage{doi}
\usepackage{amsmath}
\usepackage{physics}
\usepackage{amssymb}

\title{The effect of network topologies on fully decentralized learning: a preliminary investigation}

\author{ 
    \href{https://orcid.org/0009-0001-4307-2720}{\includegraphics[scale=0.06]{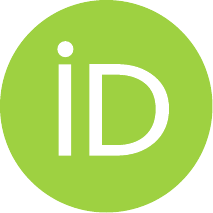}\hspace{1mm}
    Luigi ~Palmieri} \\
	IIT-CNR\\
    Pisa, Italy\\
    \texttt{luigi.palmieri@iit.cnr.it} \\
\And
    \href{https://orcid.org/0000-0001-5574-7847}{\includegraphics[scale=0.06]{orcid.pdf}\hspace{1mm}Lorenzo ~Valerio}\thanks{ Both authors contributed equally to this research} \\	IIT-CNR\\
    Pisa, Italy\\
    \texttt{lorenzo.valerio@iit.cnr.it}\\
\AND
    \href{https://orcid.org/0000-0001-5080-8110}{\includegraphics[scale=0.06]{orcid.pdf}\hspace{1mm}Chiara ~Boldrini}$^*$\\	IIT-CNR\\
    Pisa, Italy\\
    \texttt{chiara.boldrini@iit.cnr.it}\\
\And
\href{https://orcid.org/0000-0002-1694-612X}{\includegraphics[scale=0.06]{orcid.pdf}\hspace{1mm}Andrea ~Passarella}\\	IIT-CNR\\
    Pisa, Italy\\
    \texttt{andrea.passarella@iit.cnr.it}\\
}

\date{}

\renewcommand{\headeright}{}
\renewcommand{\undertitle}{}

\hypersetup{
pdftitle={The effect of network topologies on fully decentralized learning: a preliminary investigation},
pdfsubject={cs.AI, cs.IT},
pdfauthor={Luigi ~Palmieri, Lorenzo ~Valerio, Chiara ~Boldrini, Andrea ~Passarella},
pdfkeywords={Decentralized Learning, Graph topologies, non-IID data},
}

\begin{document}
\maketitle

\begin{abstract}
In a decentralized machine learning system, data is typically partitioned among multiple devices or nodes, each of which trains a local model using its own data. These local models are then shared and combined to create a global model that can make accurate predictions on new data. In this paper, we start exploring the role of the network topology connecting nodes on the performance of a Machine Learning model trained through direct collaboration between nodes. We investigate how different types of topologies impact the "spreading of knowledge", i.e., the ability of nodes to incorporate in their local model the knowledge derived by learning patterns in data available in other nodes across the networks. Specifically, we highlight the different roles in this process of more or less connected nodes (hubs and leaves), as well as that of macroscopic network properties (primarily, degree distribution and modularity). Among others, we show that, while it is known that even weak connectivity among network components is sufficient for \emph{information} spread, it may \emph{not} be sufficient for \emph{knowledge} spread. More intuitively, we also find that hubs have a more significant role than leaves in spreading knowledge, although this manifests itself not only for heavy-tailed distributions but also when ``hubs" have only moderately more connections than leaves. Finally, we show that tightly knit communities severely hinder knowledge spread.
\end{abstract}


\section{Introduction}
\label{sec:intro}

We are witnessing a paradigm shift from centralised AI systems to decentralised ones, motivated by the fact that 
data generators at the edge of the network are increasingly less inclined to share their private data with third parties, even under the promise of gaining an advantage from centralized AI-based services. 
Solutions based on decentralised AI systems, such as Federated Learning (FL)~\cite{mcmahan_communication-efficient_2017}, are promising candidates to address these concerns. In FL, the idea is to let the edge devices keep the data locally and have a central role in the knowledge extraction process. Specifically, the devices collaboratively train an AI model without sharing any raw data with each other. The only information shared is the parameters of the AI models that are trained locally. These parameters are subsequently aggregated to form a better and capable \emph{global} model that is iteratively refined through successive rounds of collaboration. 

The standard definition of the FL framework assumes a starred network topology, i.e., a central parameter server oversees the entire process, coordinating the operations of many clients. However, the presence of a centralised controller might also represent a single point of failure, a potential bottleneck when the number of devices collaborating scales up to millions of devices, and an obstacle to spontaneous, direct collaborations among users. 
Decentralised Federated Learning (DFL) represents an alternative to centralised Federated Learning. In DFL the connectivity between devices is represented by a generic graph, and the devices involved in the learning process typically collaborate only with their neighbours. 
The lack of a central controller overcomes the single point of failure problem but introduces other aspects that need to be investigated. Specifically, we claim that, in this scenario, the information locality and the network topology strongly affect the dynamics of the learning process, i.e. how fast and effective the spreading of knowledge about the class labels is.  

While previous work~\cite{sun_decentralized_2023} assumes that the network topology can be controlled by the network operator and optimized to make the learning process more efficient or scalable, we argue that complete decentralization can only be achieved by letting user devices spontaneously organize themselves. This implies that the network topology, in these settings, cannot be controlled by the operator. For example, an edge between two nodes in the graph may represent a trust relationship or willingness to cooperate. If the edges are weighted, the strength of the weights expresses the intensity of trust or cooperation. With this approach, users are free to cooperate with whomever they want, and the operator has no control over the cooperation patterns. Although this scenario poses a challenge from a learning perspective, it also fully exploits the human-centric, impromptu potential of fully decentralized learning systems.

In light of these considerations, the research question we tackle in this paper is how the network topology affects the learning process in a fully decentralized learning system. Specifically, we consider a scenario where a set of devices that are connected through a complex network topology collaborate to train a \emph{common} AI model in a completely decentralised fashion. According to the DFL framework, each device in the system receives a set of models from its neighbours in the graph. These models are first aggregated with the local one (typically through a weighted average) to get a refreshed aggregate model, which is further updated through a number of local training epochs (on local data). Finally, the newly updated models are shared between the neighbours. The devices repeat this until a stopping condition is met.

In our paper, we consider three topologies: Erdos-Renyi, Barabasi-Albert, and Stochastic Block Model, and we analyse through simulations how the learning process is affected by them, considering different non-IID data partitioning settings. The main take-home messages are the following: (i) the initial data distribution on high vs low degree nodes plays a key role in the final accuracy of a decentralized learning process, (ii) when high-degree nodes possess more knowledge, such knowledge spreads  easily in the network, (iii) vice versa, when low-degree nodes have more knowledge, knowledge spreads better when the network is less connected (at first counterintuitive, but connectivity dilutes knowledge in average-based decentralized learning), (iv) when users are grouped in tightly knit communities, it is very difficult for knowledge to circulate outside of the community.
\vspace{-10pt}
\section{Related work}
\label{sec:rel_work}

DFL extends the typical settings of FL, e.g., data heterogeneity and non-convex optimisation, by removing the existence of the central parameter server. 
This is a relatively new topic that is gaining attention from the community since it fuses the privacy-related advantages of FL with the potentialities of decentralised and uncoordinated optimisation and learning. 
In \cite{Roy2019}, the authors define a DFL framework for a medical application where a number of hospitals collaborate to train a Neural Network model on local and private data. In \cite{Lalitha2019} the authors propose a Bayesian-like approach where the aggregation phase is done by minimising the Kullback-Leibler divergence between the local model and the ones received from the peers. All these approaches are still considering that the nodes perform just one local update before sharing the parameters (or gradients) with the peers in the network. This aspect is relaxed in \cite{savazzi_federated_2020} and \cite{sun_decentralized_2023}. In \cite{savazzi_federated_2020}, the authors propose a federated consensus algorithm extending FedAvg from~\cite{mcmahan_communication-efficient_2017} in decentralised settings, mainly considering industrial and IoT applications. The authors of \cite{sun_decentralized_2023} propose a Federated Decentralised Average based on SGD where the authors include a momentum term to counterbalance the possible drift introduced by the multiple updates and a quantization scheme to reduce communications. 

Most of these papers assume a decentralised system made of a few nodes connected through controlled network topologies, e.g., rings and full meshes. In this paper, we start exploring how the network topology affects the learning process of different decentralised learning schemes under data heterogeneity. To the best of our knowledge, this is the first paper that considers complex network topologies and relates their key features to the performance of  decentralised learning.

%

\section{Decentralized learning}
\label{sec:system_model}

We represent the network connecting the nodes as $\mathcal{G}(\mathcal{V},\mathcal{E})$, where $\mathcal{V}$ denotes the set of nodes and $\mathcal{E}$ the set of edges. Without loss of generality, we assume that the graph represents a social network. Exactly the same concepts can be translated to different application domains. We denote with $\omega_{ij}$ the weights on the edge between nodes $i$ and $j$ which, in the case of a social network, would represent the trust/social intimacy between the two nodes. The self-trust $\omega_{ii}$ is a pseudo-parameter with which we capture the importance placed by node $i$ on itself. We assume that only nodes sharing an edge are willing to collaborate with each other: effectively, we use the existence of a social relationship as a proxy of trust.

Each node $i \in \mathcal{V}$ is equipped with a local training dataset~$\mathcal{D}_i$ (containing tuples of features and labels $(x,y) \in \mathcal{X} \times \mathcal{Y}$) and a local model $h_i$ defined by weights $\mathbf{w}_i$, such that $h_i(\mathbf{x}; \mathbf{w}_i)$ yields the prediction of label $y$ for input $\*x$. Let us denote with $\mathcal{D} = \bigcup_i \mathcal{D}_i$ and with $\mathcal{P}$ the label distribution in $\mathcal{D}$. In general, $\mathcal{P}_i$ (i.e., the label distribution of the local dataset on node $i$) may be different from $\mathcal{P}$. This captures a realistic non-IID data distribution. 
At time 0, the model $h(\cdot; \*w_i)$ is, as usual, trained on local data, by minimizing a target loss function $\ell$ -- i.e., $\*w_i = \mathrm{argmin}_{\*w} \sum_{k = 1}^{|\mathcal{D}_i |} \ell(y_k, \*w \*x_k)$, with $(y_k, \*x_k) \in \mathcal{D}_i$. 
%
%
%
We assume that nodes entertain a certain number of communication rounds, where they exchange and combine local models. At each communication round, a given node receives the local models from its neighbors in the social graph, and averages it with its local model. 
Specifically, at each step $t$, the local model of the given node and the local models from the node's neighbors are averaged as follows:
\begin{equation}
    \*w_i(t) \leftarrow \frac{\sum_{j \in \mathcal{N}(i)} \omega_{ij} \alpha_{ij} \*w_j(t-1)}{\sum_{j \in \mathcal{N}(i)} \omega_{ij}},
\end{equation}
where we have denoted with $\mathcal{N}(i)$ the neighborhood of node~$i$ including itself and $\alpha_{ij}$ is equal to $\frac{|\mathcal{P}_j|}{\sum_{j \in \mathcal{N}_i} | \mathcal{P}_j|}$ (and captures the relative weight of the local dataset of node $j$ in the neighborhood of node $i$).
Once the aggregation of models is performed, the local model is trained again on the local data (in this paper, we use learning rate $\eta$ and momentum $\mu$).

This strategy (hereafter, DecAvg) is the natural extension of FedAvg (the most well-known FL approach~\cite{mcmahan_communication-efficient_2017}) to a decentralized setting and is a social-aware generalization of similar strategies proposed in~\cite{sun_decentralized_2023,savazzi_federated_2020}: the aggregation is performed not by the central controller (as in federated settings) but by each node, whereby each node averages the model received from a given neighbor based on the strength of the social link connecting them.

\vspace{-5pt}
\section{Network topologies}
\label{sec:netowork_topologies}

In this paper, we analyze three different topologies to investigate how their different properties impact knowledge diffusion in a fully decentralized learning process: Erdos-Renyi (ER), Barabasi-Albert (BA), and  Stochastic Block Model (SBM) graphs.

The ER model is a model for generating random graphs with a homogeneous structure, where nodes are connected to each other with a fixed probability. 
ER is defined by two parameters: $N$, the number of nodes in the network, and $p$, the probability of an edge existing between any two nodes in the network (regardless of their degree). The ER model shows a phase transition when the fixed probability $p$ approaches the critical value $p^*=\ln(N)/N $~\cite{erdHos1960evolution}. Specifically, the value $p^*$ is a sharp threshold for the connectedness of the network: for values of $p$ above $p^*$ the network goes from being mostly disconnected to showing a growing clustering coefficient. 

The BA is an algorithm for generating random scale-free networks, i.e., networks with a power-law (or scale-free) degree distribution, using a preferential attachment mechanism~\cite{albert2002statistical}. In the BA model, nodes are connected preferentially based on their degree. Specifically, the probability of an edge forming between two nodes is proportional to the nodes' degree, which leads to the emergence of a scale-free degree distribution. Since the degree distribution follows a power law, few nodes have a very high degree while most nodes have a low degree. This can result in a structure with few well-connected hubs, which is known to facilitate information flow across the network. A BA network is defined by two parameters: $N$, the number of nodes in the network, and $m$, the number of edges added to the network for each new node (hence, the minimum degree of nodes). 

The SBM is a probabilistic model for networks that exhibit a modular structure, i.e., the SBM generates a network with a clear community structure where nodes are grouped together based on their connectivity patterns~\cite{lee2019review}.
Nodes belonging to the same group are more closely connected to each other than to nodes in another group.
Formally, the SBM is defined by the following parameters: $N$, the number of nodes in the network;  $B$, the number of communities (called blocks); ${n_1, n_2, ..., n_B}$, the sizes of the blocks where $n_i$ is the number of nodes in block $i$; $p_{ij}$, the probability of an edge existing between a node in block $i$ and a node in block $j$ (with $p_{ii}$ the probability of links inside the block).

These three models each capture important properties of complex networks. ER graphs are homogeneous in terms of degree and with a low clustering coefficient. BA graphs are characterised by a very skewed degree distribution with few high-degree nodes and many low-degree nodes. Finally, SBM graphs feature a well-defined community structure.

\section{Evaluation}

\subsection{Experimental settings}
\label{sub:settings}

In this paper, we consider unweighted graphs with 100 nodes, where edges are generated as follows. 

\textbf{ER.} Three different conditions regarding the parameter $p$ are analyzed: just below the critical value ($p \lessapprox p^*$), at the critical value ($p = p^*$) and just above the critical value ($p \gtrapprox p^*$). 

\textbf{BA.} Three different cases regarding the parameter of preferential attachment are chosen: $m = 2,5,10$, leading to networks with increasingly higher node degree.

\textbf{SBM.} Nodes are grouped into 4 communities of equal size (25 nodes each). The probability of extrinsic connections $(p_{ij}, \; j \neq i)$ is set to~0.01, whereas the probability of intrinsic connections $(p_{ii})$ is set to 0.8 in one case study and 0.5 in the second case study.

For our experiments, we choose the widely used MNIST image dataset. This dataset contains a set of handwritten digits, thus data are divided into 10 classes. The goal of the analysis is to characterise the effect of the network topology in the knowledge spreading process, i.e., the ability of nodes to learn data patterns they have \emph{not} seen locally, but that other nodes in the network have seen. Therefore, we  split the MNIST classes across nodes as follows. Note that, on the assigned classes, each node gets the same amount of images.

For ER and BA networks, we divide the 10 MNIST classes into two groups: the first group (G1) is composed of classes 0, 1, 2, 3, 4, the second group (G2) of classes 5, 6, 7, 8, 9. All nodes receive an equal share (selected randomly) of data from G1. Data from G2 are allocated only to subsets of nodes. Specifically, we consider two different cases, whereby data in G2 are allocated to the 10\% highest-degree and lowest-degree nodes, respectively. The rationale is thus to allocate ``full knowledge" (i.e., a complete subset of all classes) either to high-degree or to low-degree nodes, and study the effect of the network topology in both cases. In the following, these configurations are referred to as ``hub-focused" and ``edge-focused", respectively. Specifically, starting from the node(s) with the highest (lowest) degree, we pick nodes until we reach 10\% of the network. In case adding all nodes at a given degree results in more than 10\% of the network, we randomly pick, among nodes with that degree, a subset that allows us to fill the 10\% subset.

For SBM networks, we divide the dataset classes into subsets based on the communities the nodes belong to, without overlap. Therefore, since we study  SBM topologies with 4 communities, each community gets two classes: community 1 sees classes 0 and 1; community 2 sees classes 2 and 3; community 3 sees classes 4 and 5; community 4 sees classes 6 and 7. This data distribution is designed to challenge the knowledge spreading process, since maximum learning accuracy can only be achieved if information from all the external communities is brought into the local one. 

For the learning task, we consider a simple classifier as the learned model and focus on two performance figures. On the one hand, we consider the accuracy over time at each node, to assess the effectiveness and speed of knowledge diffusion across the network. On the other hand, for SBM networks, we also investigated the average confusion matrix across nodes of the same community. Specifically, for each node we compute the confusion matrix for the MNIST classes, and then take the average across all nodes in the same community.

We implemented the DecAvg scheme within the custom SAISim simulator, available on Zenodo\footnote{\url{https://doi.org/10.5281/zenodo.5780042}}. SAISim is developed in Python and leverages state-of-the-art libraries such as PyTorch and NetworkX for deep learning and complex networks, respectively. On top of that, SAISim implements the primitives for supporting fully decentralized learning. The local models of nodes are Multilayer Perceptrons with three layers (sizes 512, 256, 128) and ReLu activation functions. SGD is used for the optimization, with learning rate~$0.001$ and momentum~$0.5$. 

\subsection{Results}
\label{sec:results}


\subsubsection{ER}
\label{sec:results_er}
\begin{figure*}[h!]
    \centering
    \includegraphics[width = \textwidth]{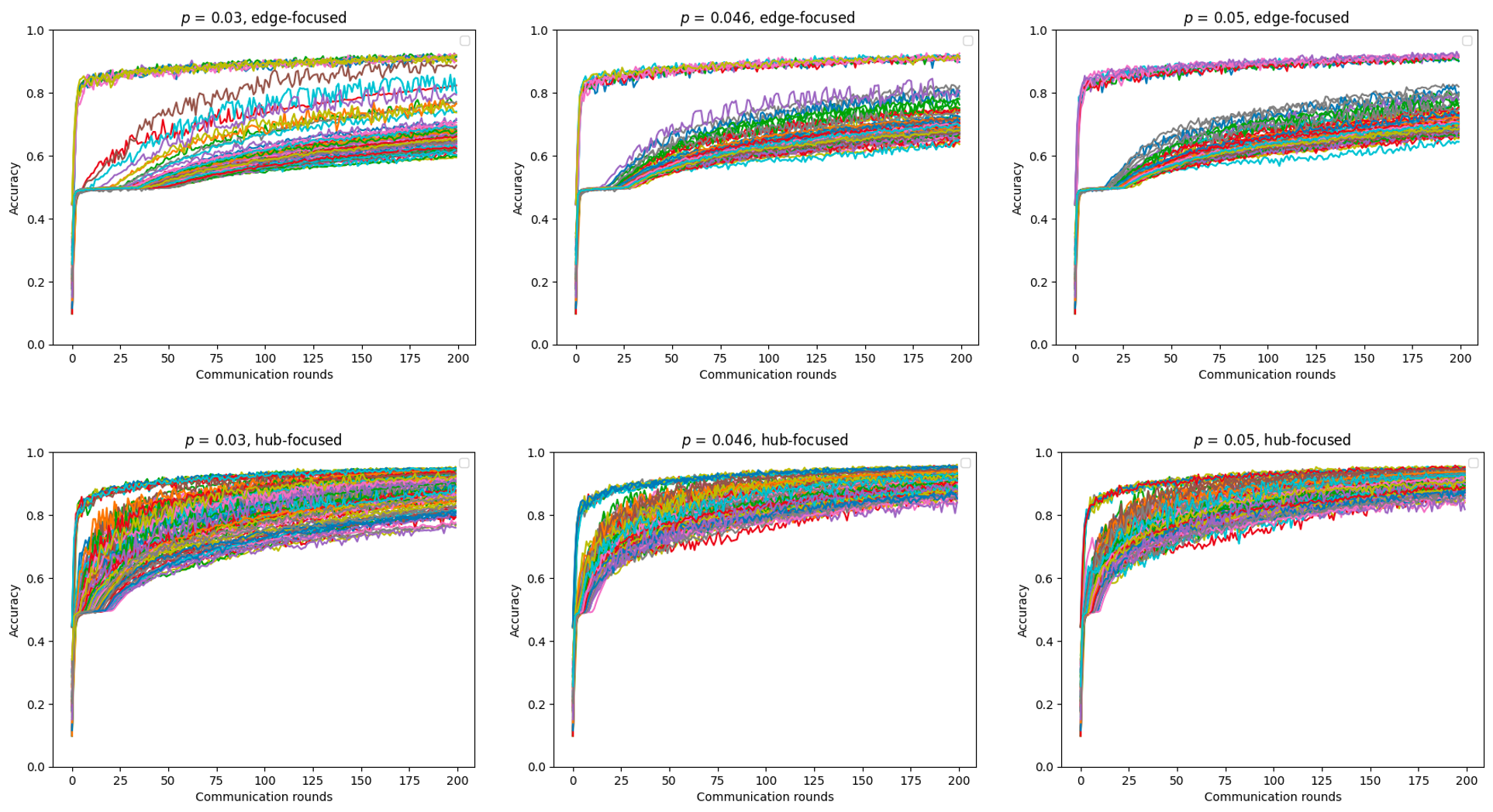}
    \caption{Accuracy in ER networks (all nodes). From left to right: increasing values of $p$; from top to bottom: edge-focused scenario and hub-focused scenario.} 
    \label{fig:er_accuracy_tot}
\end{figure*}

\begin{figure}[h!]
    \centering
    \includegraphics[width = \textwidth]{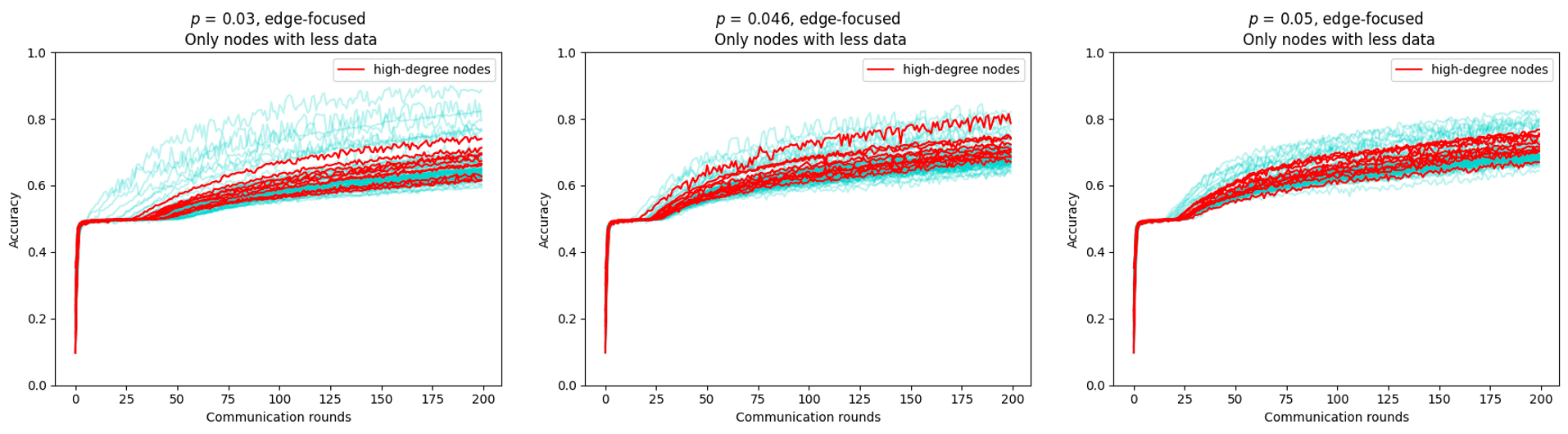}
    \caption{Accuracy in ER networks (nodes with only G1 images, edge-focused scenario), curves of the high-degree nodes in red.}
    \label{fig:er_accuracy_lessdata_hubshighlight}
\end{figure}

As explained in Section~\ref{sub:settings}, for the ER model we want to evaluate scenarios around the critical value of connectedness. Considering our settings, the critical value $p^*$ is 0.046, hence the chosen values of $p$ are $0.03$, $0.046$, $0.05$. 
In Figure~\ref{fig:er_accuracy_tot} we show the evolution of the accuracy per node over time (each curve corresponds to one node). The top-row plots refer to the edge-focused case, where the digits in class G2 are assigned to leaf nodes, while the bottom-row plots correspond to the hub-focused scenario, where the well-connected nodes get all the images in G2. The first interesting result is that, in the edge-focused case, the separation between the curves of the nodes having more data and those that have fewer data is more evident. This means that when leaf nodes possess the missing classes (G2 images), it is harder for knowledge to circulate.
This finding holds when we vary $p$ (which increases from left to right). However, in the edge-focused scenario with below-critical $p$ (top left corner), we see that some nodes are eventually able to reach the accuracy of the nodes holding more data. But as we go from left to right, i.e., from less to more connected networks, this ability vanishes. This is quite interesting as well as counterintuitive. When the network is more connected, high-degree nodes (we refrain from calling them hubs, as in ER networks hubs do not exist) get proportionally more connected than leaf nodes. As G2 data are present only in leaf nodes, the additional connectivity plays against knowledge diffusion from leaves, as leaves' models are averaged among larger sets of nodes (due to higher degrees of more connected nodes), and therefore their knowledge ``weighs" less in the average. This is also clear in Figure~\ref{fig:er_accuracy_lessdata_hubshighlight}, where we show the accuracy only for nodes that are only assigned G1 images, and highlight the curves of the highest-degree nodes, which remain at low accuracy. The more the network is connected, the more high-degree nodes ``drag" other nodes toward their performance. In the hub-focused case (lower part of Figure~\ref{fig:er_accuracy_tot}) results are more intuitive, as the highest degree nodes enjoy higher accuracy, and are able to drag all the other nodes closer and closer to their performance.

\begin{figure}[h!]
    \centering
    \includegraphics[width = \textwidth]{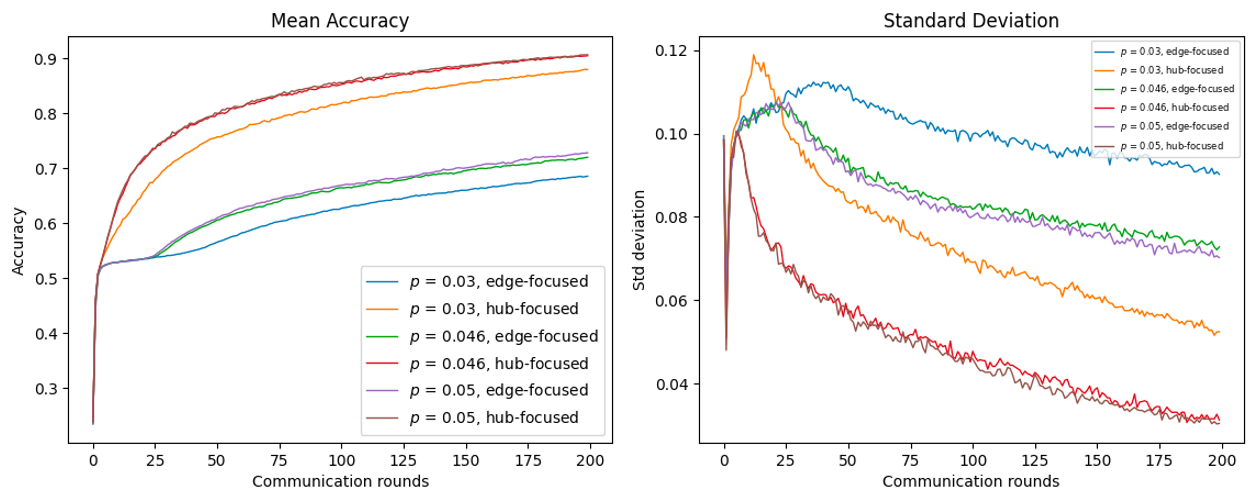}
    \caption{ER (all nodes): accuracy mean and std.}
    \label{fig:er_mean_acc_std_all}
\end{figure}

In Figure~\ref{fig:er_mean_acc_std_all}, we aggregate the accuracy among all nodes in the same experiment and we show the evolution over time of the average and standard deviation of the accuracy. 
Coherently with Figure~\ref{fig:er_accuracy_tot}, the average accuracy in the edge-focus case is lower than in the hub-focus case. This is again due to high-degree nodes (in the former case) blocking the spreading of knowledge from leaves.
An interesting feature can be highlighted by comparing, in both cases, the standard deviation curves for networks below the critical threshold ($p = 0.03$, blue and orange curves in the figure) and the rest, which clearly show a lower rate of decay (after an initial increase). This can be attributed to the effect of longer paths when $p = 0.03$ (because the network is less connected), which hinders the alignment of local models across nodes, which are in this case farther away than for more connected networks (values of $p$ at and beyond the critical threshold). 
Also, the standard deviation values are generally higher in the edge-focus case, as hubs hinder other nodes to incorporate knowledge from leaves. This effect, again, is less pronounced for less connected networks, where some nodes are able to ``escape" the dragging effect of high-degree nodes, thus resulting in higher accuracy (but higher standard deviation) at the level of the entire network.

Finally, we can observe another overall effect by jointly analysing the accuracy per node, the average accuracy and the standard deviation for the edge-focus case. Specifically, \emph{on average} nodes in the least connected network achieve lower accuracy (Figure~\ref{fig:er_mean_acc_std_all}, orange and blue curves) even though, as a side effect of the higher standard deviation, some non-leaf nodes are able to increase their accuracy escaping the attraction effect of high-degree nodes. As the network becomes more and more connected, no node in the ``non-leaf and non-hub" class is able to reach the same accuracy, even though, on average, the overall accuracy increases. This means that, as a side effect of additional connectivity, hubs tend to increase their accuracy, even though they block other nodes to efficiently incorporate models of leaf nodes.



\subsubsection{BA}
\label{sec:results_ba}

\begin{figure}[h]
    \centering
    \includegraphics[width =\textwidth]{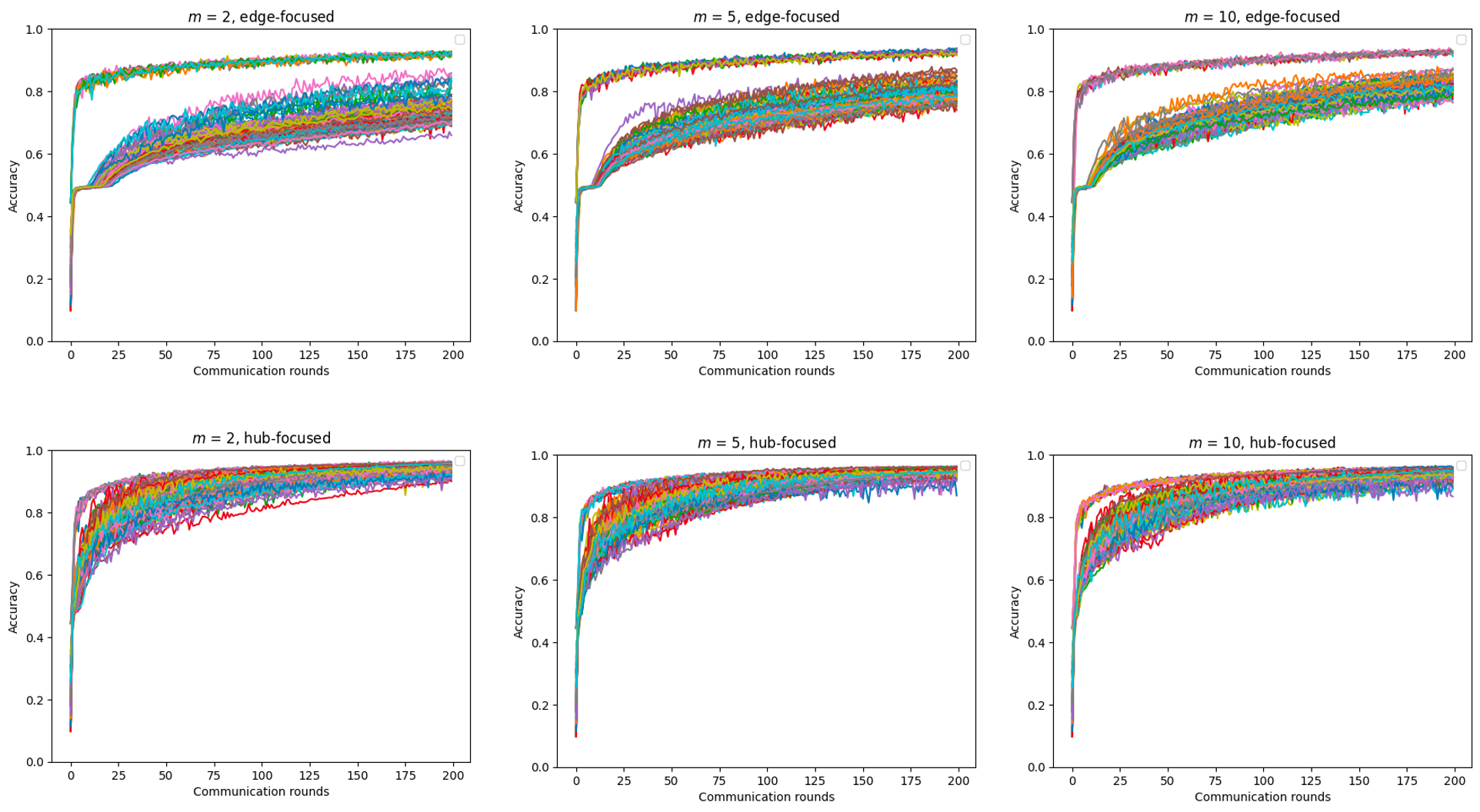}
    \caption{Accuracy in BA networks (all nodes). From left to right, the parameter $m$ increases; from top to bottom: edge-focused and hub-focused scenario.} 
    \label{fig:ba_accuracy_tot}
\end{figure}

\begin{figure}[h]
    \centering
    \includegraphics[width=\textwidth]{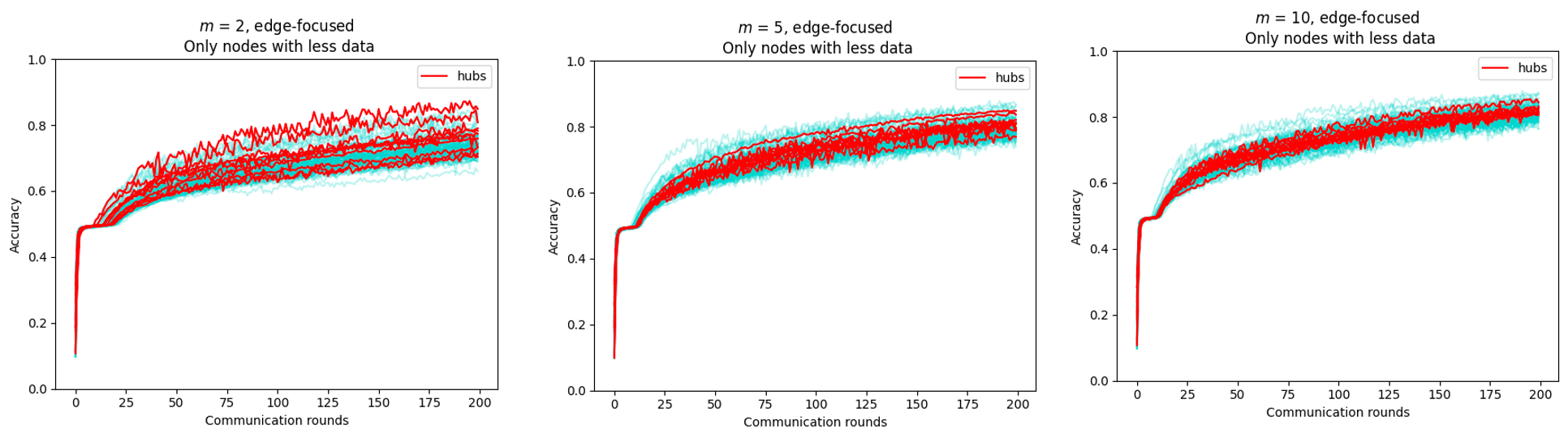} 
    \caption{Accuracy in BA networks (nodes with only G1 images, edge-focused scenario), high-degree nodes in red.} 
    \label{fig:ba_accuracy_lessdata_hubshighlight}
\end{figure}

\begin{figure}[h]
    \centering
    \includegraphics[width=\textwidth]{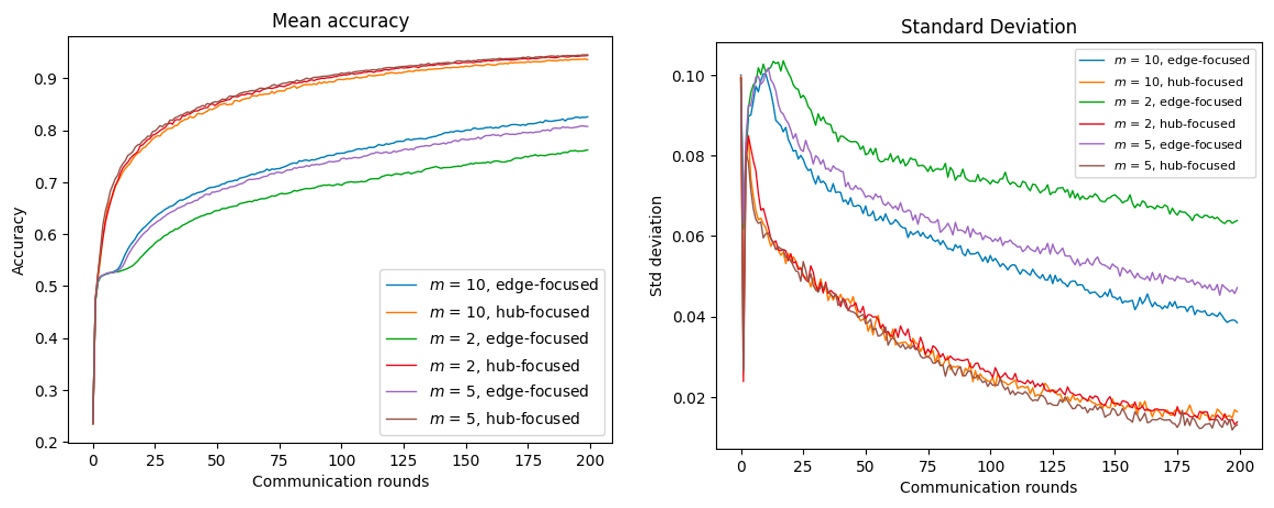}
    \caption{BA (all nodes): accuracy mean and std.} 
    \label{fig:ba_mean_acc_std_all}
\end{figure}

As discussed earlier, for the Barabasi-Albert model we analyze three different settings varying the parameter related to the preferential attachment: $m \, = \,2,5,10$.
We show the accuracy over time in Figure~\ref{fig:ba_accuracy_tot}. First, in the hub-focused case (bottom row), the performance for varying values of the minimum degree $m$ is basically indistinguishable (this is confirmed by looking at the average and standard deviation of the accuracy in Figure~\ref{fig:ba_mean_acc_std_all}). This means that hubs (in this case, high-degree nodes are \emph{real} hubs) spread knowledge extremely efficiently, irrespective of the connectivity of the rest of the nodes.
%
%
The edge-focused case (top row of Figure~\ref{fig:ba_accuracy_tot}) is more interesting. As was the case with ER networks, leaf nodes are not able to spread their knowledge efficiently and the accuracy gap between leaf nodes and non-leaf nodes remains strong all throughout. In Figure~\ref{fig:ba_mean_acc_std_all}, we observe that larger values of $m$ (i.e., stronger connectivity) help improve the average accuracy but not significantly, while the variability is reduced. 
The hubs in the edge-focused scenario seem to benefit from the existence of lower-degree nodes, corresponding to smaller $m$ (Figure~\ref{fig:ba_accuracy_lessdata_hubshighlight}). Vice versa, when the degree of the other nodes increases, their accuracy is dragged down by them. This results, though, in better average accuracy and reduced variability.


\subsubsection{SBM}
\label{sec:results_sbm}

\begin{figure}[h!]
    \centering
    \includegraphics[width = \textwidth]{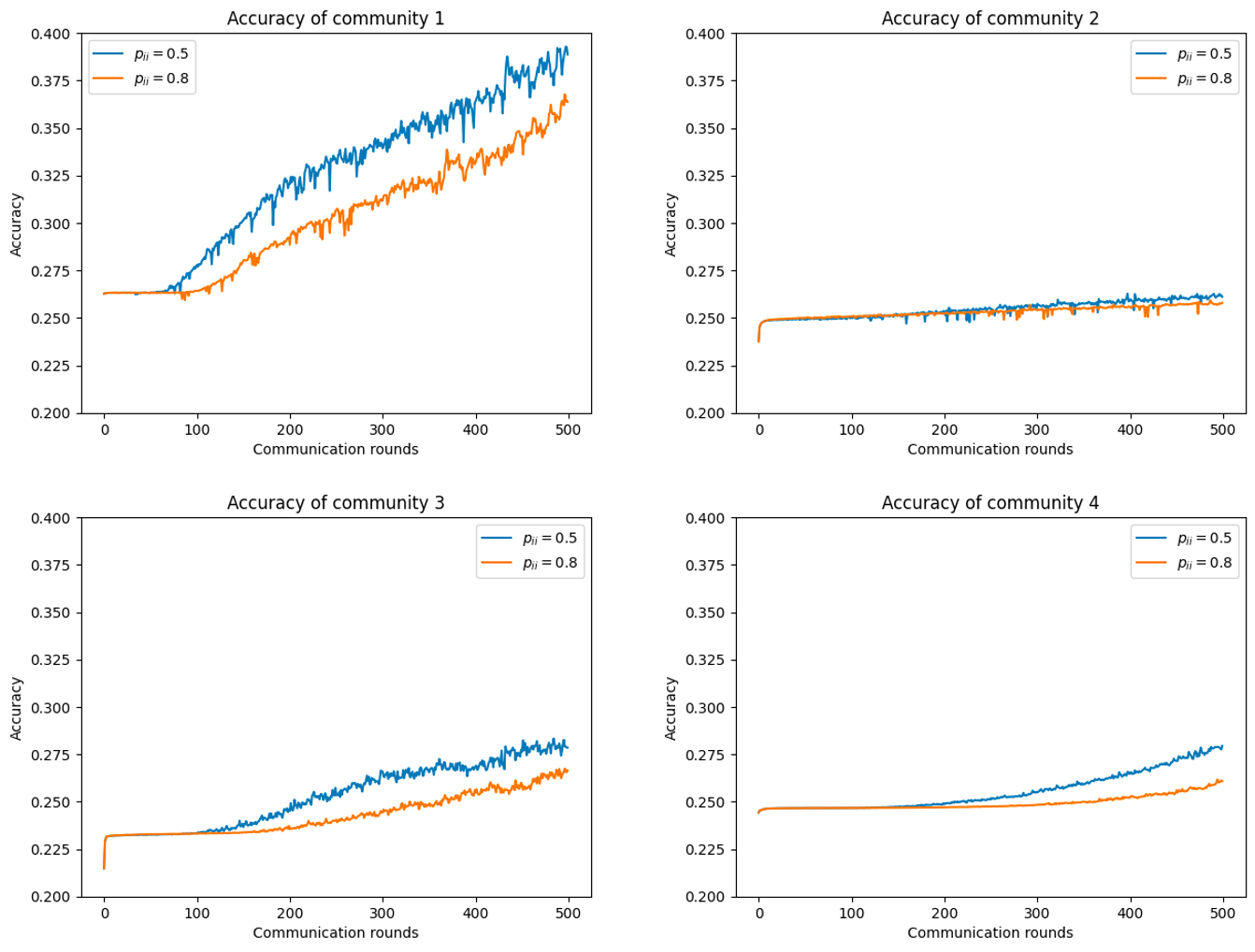}
    \caption{Mean accuracy in SBM communities.} 
    \label{fig:sbm_mean_acc_per_community}
\end{figure}

The SBM topology is different from the previous two as it features four clearly separated communities, with sporadic intercommunity links. For the intracommunity connectivity, we test two scenarios (lower and higher intracommunity connectivity, corresponding to $p_{ii}= 0.5$ and $p_{ii}= 0.8$). Recall that each community holds two non-overlapping MNIST classes (hence, classes 8 and 9 are discarded). Using only intracommunity information, nodes can at most achieve a 0.25 accuracy (perfect classification of the two classes in their training data, zero knowledge on the other six). In order to go beyond 0.25, knowledge must be circulated across communities. The question is whether the occasional intercommunity edges are enough for that, and if the densely connected communities will make it harder for external information to percolate internally. Figure~\ref{fig:sbm_mean_acc_per_community} shows that the latter is happening: with $p_{ii}=0.5$ (less dense communities) the average accuracy grows faster than with $p_{ii}=0.8$. 
The figure also reveals the presence of stragglers, for whom catching up with the rest of the network takes some time. Interestingly, entire communities appear to be stragglers. Table~\ref{tab:confmat_05} shows that communities are, as expected, very good at classifying classes they have in their local training data. Vice versa, it is very hard for external knowledge to enter the communities. Table~\ref{tab:confmat_05} also shows the number of links pointing towards external communities, which are the conduit for knowledge diffusion. Community 2 enjoys fewer external links, and indeed its learning process is very slow and mediocre (Figure~\ref{fig:sbm_mean_acc_per_community}). However, Community 1 has only slightly fewer links than Community 3, but its learning process is faster and more accurate. We argue that the specific classes assigned to communities might play a role in this case, and we will investigate this aspect in future work.

\begin{table}[h!]
   \footnotesize
   \centering
   \begin{tabular}{@{}lllll@{}}
   \toprule
       \textbf{Class} & \textbf{Comm. 1} & \textbf{Comm. 2} & \textbf{Comm. 3} & \textbf{Comm. 4} \\
       & (-, 5,9,7) & (5,-,7,3) & (9,7,-,8) & (7,3,8,-)\\\midrule
       0 & 0.9961 & 0.0002 & 0.0004 & 0.0043 \\ 
       1 & 0.9992 & 4e-05 & 0.0684 & 0.0079 \\ \hline
       2 & 0.0 & 0.9868 & 4e-05 & 0 \\ 
       3 & 0.0146 & 0.9802 & 0 & 0.0002 \\ \hline
       4 & 0.1824 & 0.0076 & 0.9971 & 0.0006 \\ 
       5 & 0.0011 & 4e-05 & 0.9972 & 0.0011 \\ \hline
       6 & 0.0039 & 0 & 0.0003 & 0.9979 \\
       7 & 0.272 & 0.0101 & 0.0225 & 0.9966 \\ \bottomrule \\
   \end{tabular} 
\caption{Accuracy per MNIST class and community for SBM with $p_{ii} \, = \, 0.5$. For each community, we report in square brackets the number of edges pointing toward external community 1, 2, 3, 4, respectively.} 
\label{tab:confmat_05}
\end{table}

\clearpage
\section*{Acknowledgments}
\thanks{This work was partially supported by the H2020 HumaneAI Net (952026) and by the CHIST-ERA-19-XAI010 SAI projects. C. Boldrini's work was partly funded by the PNRR - M4C2 - Investimento 1.3, Partenariato Esteso PE00000013 - "FAIR", A. Passarella's work was partly funded by the PNRR - M4C2 - Investimento 1.3, Partenariato Esteso PE00000001 - "RESTART", both funded by the European Commission under the NextGeneration EU programme.}

\bibliographystyle{abbrv}
\bibliography{sai} 

\begin{thebibliography}{1}

\bibitem{albert2002statistical}
R.~Albert and A.-L. Barab{\'a}si.
\newblock Statistical mechanics of complex networks.
\newblock {\em {Rev. Mod. Phys.}}, 74(1):47, 2002.

\bibitem{erdHos1960evolution}
P.~Erd{\H{o}}s, A.~R{\'e}nyi, et~al.
\newblock On the evolution of random graphs.
\newblock {\em Publ. Math. Inst. Hung. Acad. Sci}, 5(1):17--60, 1960.

\bibitem{Lalitha2019}
A.~Lalitha, X.~Wang, O.~Kilinc, Y.~Lu, T.~Javidi, and F.~Koushanfar.
\newblock Decentralized bayesian learning over graphs.
\newblock {\em arXiv}, 2019.

\bibitem{lee2019review}
C.~Lee and D.~J. Wilkinson.
\newblock A review of stochastic block models and extensions for graph
  clustering.
\newblock {\em Appl. Netw. Sci.}, 4(1):1--50, 2019.

\bibitem{mcmahan_communication-efficient_2017}
H.~B. McMahan, E.~Moore, D.~Ramage, S.~Hampson, and B.~Ag\"uera~y Arcas.
\newblock Communication-efficient learning of deep networks from decentralized
  data.
\newblock In {\em AISTATS'17}, 2017.

\bibitem{Roy2019}
A.~G. Roy, S.~Siddiqui, S.~P{\"o}lsterl, N.~Navab, and C.~Wachinger.
\newblock {BrainTorrent}: {A} {Peer}-to-{Peer} {Environment} for
  {Decentralized} {Federated} {Learning}.
\newblock {\em arXiv}, pages 1--9, 2019.

\bibitem{savazzi_federated_2020}
S.~Savazzi, M.~Nicoli, and V.~Rampa.
\newblock Federated {Learning} {With} {Cooperating} {Devices}: {A} {Consensus}
  {Approach} for {Massive} {IoT} {Networks}.
\newblock {\em IEEE Internet of Things Journal}, 7(5):4641--4654, 2020.

\bibitem{sun_decentralized_2023}
T.~Sun, D.~Li, and B.~Wang.
\newblock Decentralized federated averaging.
\newblock {\em IEEE Trans. Pattern Anal. Mach. Intell.}, 45(04):4289--4301,
  2023.

\end{thebibliography}
\end{document}